\documentclass[runningheads]{llncs}

\usepackage[mobile]{eccv}

\usepackage{eccvabbrv}

\usepackage{graphicx}
\usepackage{booktabs}

\usepackage{bm}
\usepackage{algorithm}
\usepackage{algpseudocode}
\usepackage{paralist}
\usepackage{xcolor}
\usepackage{subcaption}

\usepackage[accsupp]{axessibility}  

\usepackage[pagebackref]{hyperref}
\hypersetup{
  pdftitle={eBIRD: Event-based Intensity Image Reconstruction Using Controllable Diffusion Models},
  pdfauthor={I. Bugueno-Cordova, F. Valderrama, R. Verschae},
  pdfsubject={Computer Vision and Pattern Recognition (cs.CV)},%
  pdfkeywords={Event Cameras, Image Reconstruction, Diffusion Models, Class-Agnostic Learning, Class-Conditional Generation, Neuromorphic Vision},%
}

\usepackage{orcidlink} 

\newcommand{\method}{\textsc{eBIRD}\xspace}
\newcommand{\methodg}{\textsc{eBIRD-G}\xspace}
\newcommand{\methods}{\textsc{eBIRD-S}\xspace}

\begin{document}

\title{eBIRD: Event-based Intensity Image Reconstruction Using Controllable Diffusion Models} 

\titlerunning{eBIRD}

\author{
Ignacio Bugueno-Cordova\inst{1}\orcidlink{0000-0002-0133-0330} \and
Fabian Valderrama\inst{1} \and
Rodrigo Verschae\inst{2}\orcidlink{0000-0002-1661-3309}
}

\authorrunning{I. Bugueno-Cordova et al.}


\institute{
Institute of Engineering Sciences, Universidad de O'Higgins,
Rancagua, Chile\\
\and
The Iniciativa de Datos e Inteligencia Artificial,
University of Chile, Santiago, Chile\\
\email{i.bugueno@ieee.org}, 
\email{fabian.valderrama@ieee.org},
\email{rodrigo@verschae.org}
}

\maketitle

\begin{abstract}
    
    Intensity-image reconstruction from event streams remains a challenging problem due to the binary, sparse, and asynchronous nature of event data. 
    This work proposes \method, an event-guided reconstruction framework that combines a DDPM with ControlNet-based conditioning. We analyze generic and specialized diffusion learning strategies for handwritten digit (N-MNIST) and face (RGBE-Gaze) reconstruction using 33~ms event windows.  
    On N-MNIST, the generic model achieves the best reconstruction quality (MSE 0.0052, SSIM 0.8982, PSNR 23.34~dB), whereas the specialized model performs best on RGBE-Gaze (MSE 0.0161, SSIM 0.7605, PSNR 19.08~dB). 
    These preliminary results suggest that controllable diffusion models are a promising approach for event-guided intensity-image reconstruction, while highlighting that the preferred learning strategy depends on the reconstruction domain.
    
\end{abstract}

\section{Introduction}
\label{sec:intro}

Reconstructing intensity images from event streams remains a challenging problem due to the binary, sparse, and asynchronous nature of event data~\cite{quan2024survey,chakravarthi2024recent}. Unlike conventional cameras, event cameras encode only changes in scene brightness rather than absolute intensity values~\cite{gallego-2020,zhang2025review}, requiring appearance information to be inferred during reconstruction. Recovering intensity images complements event measurements with visual appearance cues, facilitating scene interpretation and downstream vision tasks~\cite{cazzato2024,shariff-2024,chakravarthi2024recent}.

Event-to-image reconstruction methods have explored recurrent convolutional architectures, such as E2VID~\cite{rebecq2021e2vid}, and more recently transformer-based approaches, such as ET-Net~\cite{weng2021etnet}. In parallel, denoising diffusion probabilistic models (DDPMs)~\cite{ho2020denoising} have achieved remarkable performance in image synthesis and restoration tasks. Conditional diffusion frameworks, including latent diffusion models~\cite{rombach-2022} and ControlNet~\cite{zhang-2023}, further enable image generation guided by external structural information, making them attractive for event-guided image reconstruction.

In this work, we propose \method, an event-guided intensity-image reconstruction framework based on DDPMs and ControlNet conditioning. Using the proposed framework, we study of class-agnostic and class-specific learning strategies for event-guided reconstruction. Preliminary results on the N-MNIST and RGBE-Gaze datasets suggest that class-agnostic learning is more effective for handwritten digit reconstruction, whereas class-specific fine-tuning provides better face reconstructions.

The main contributions of this work are summarized as follows:
\begin{itemize}
    \item We propose \method, a diffusion-based framework for event-guided intensity-image reconstruction using DDPMs with ControlNet conditioning.
    \item We analyze the effects of class-agnostic and class-specific learning strategies.
    \item We present qualitative and quantitative evaluations on the N-MNIST and RGBE-Gaze datasets for handwritten digit and face reconstruction, respectively.
\end{itemize}
The source code and experimental protocols will be publicly released.

\section{Related work}
\label{sec:related-work}

\paragraph{Event-based Image Reconstruction.}

Event-to-image reconstruction has evolved from optimization-based approaches~\cite{bardow-2016} to recurrent deep learning methods that exploit the temporal structure of event streams~\cite{quan2024survey}. Representative architectures include E2VID~\cite{rebecq-2019}, FireNet~\cite{scheerlinck-2020}, and SPADE-E2VID~\cite{cadena-2021}, while more recent work, such as ET-Net~\cite{weng2021etnet}, explores transformer-based reconstruction. 
Although these methods have substantially advanced event-to-image reconstruction, recovering fine image details from sparse event data remains challenging~\cite{quan2024survey}, motivating the exploration of alternative generative models.

\paragraph{Diffusion Models for Image Reconstruction.}

DDPMs~\cite{ho2020denoising,dhariwal-2021,rombach-2022} have recently achieved remarkable performance in image synthesis and restoration. Conditional diffusion models, such as ControlNet~\cite{zhang-2023}, further enable image generation guided by structural information while preserving high visual fidelity. 
More recently, diffusion models have also been explored for event-guided image reconstruction~\cite{Lin-Zhu,liang2024e2vidiff}, showing promising results for recovering intensity images from event streams. Building upon these advances, this work explores diffusion-based event reconstruction under both class-agnostic and class-specific learning strategies.

\section{Preliminaries}

\subsection{Event Cameras and Event Representation}

\paragraph{Event Camera Model.}

Event cameras asynchronously measure changes in logarithmic image intensity instead of capturing complete image frames. Let
\(L(\mathbf{u},t)\doteq\log(I(\mathbf{u},t))\)
denote the logarithmic intensity at pixel
\(\mathbf{u}=(x,y)^{\mathsf{T}}\)
and time \(t\). An event
\((\mathbf{u}_k,t_k,p_k)\)
is generated whenever the brightness change reaches a contrast threshold \(C>0\):
\begin{equation}
L(\mathbf{u}_k,t_k)-L(\mathbf{u}_k,t_k-\Delta t_k)=p_kC,
\end{equation}
where \(p_k\in\{-1,+1\}\) denotes the event polarity. Over a temporal interval, the sensor outputs the event stream
$
\mathcal{E}
=
\left\{
(\mathbf{u}_k,t_k,p_k)
\right\}_{k=1}^{N},
$
which can be transformed into different representations for downstream processing~\cite{gallego-2020}.

\paragraph{Event Representation.}

Common event representations include event frames, time surfaces, voxel grids, and reconstructed intensity images~\cite{gallego-2020}. In this work, event frames are adopted, which accumulate event polarities within a temporal window \(T\):
\begin{equation}
F(x,y)
\doteq
\sum_{k:t_k\in T}
p_k
\delta_{x,x_k}
\delta_{y,y_k},
\end{equation}
where \(\delta\) denotes the Kronecker delta. These event frames serve as the conditioning input to the proposed framework.

\subsection{Denoising Diffusion Probabilistic Models}

\paragraph{Forward Process.}

DDPMs~\cite{ho2020denoising} learn to reverse a gradual noising process. During training, Gaussian noise is progressively added to a clean sample \(x_0\) according to
\begin{equation}
q(x_t|x_{t-1})
=
\mathcal{N}
(x_t;
\sqrt{1-\beta_t}\,x_{t-1},
\beta_tI),
\end{equation}
where \(x_t\) denotes the noisy sample at timestep \(t\), \(\beta_t\) is the variance schedule controlling the amount of injected noise, and \(I\) is the identity matrix. By defining
\(\alpha_t=1-\beta_t\)
and
\(\bar{\alpha}_t=\prod_{s=0}^{t}\alpha_s\),
the noisy sample at any timestep can be obtained directly as
\begin{equation}
x_t
=
\sqrt{\bar{\alpha}_t}\,x_0
+
\sqrt{1-\bar{\alpha}_t}\,\epsilon,
\qquad
\epsilon\sim\mathcal{N}(0,I).
\label{eq:forward_diffusion}
\end{equation}
where \(x_0\) is the clean image, \(\epsilon\) is Gaussian noise, and \(\bar{\alpha}_t\) controls the signal-to-noise ratio at timestep \(t\).

\paragraph{Reverse Process.}

The reverse process reconstructs the clean sample from
\(x_T\sim\mathcal{N}(0,I)\)
through
\begin{equation}
p_\theta(x_{t-1}|x_t)
=
\mathcal{N}
(x_{t-1};
\mu_\theta(x_t,t),
\Sigma_\theta(x_t,t)).
\end{equation}
Here, \(\mu_\theta(\cdot)\) and \(\Sigma_\theta(\cdot)\) denote the learned mean and covariance of the reverse transition, parameterized by the network weights \(\theta\). Rather than predicting the clean image directly, the model estimates the injected Gaussian noise
\(\epsilon_\theta(x_t,t)\),
leading to the reverse transition
\begin{equation}
x_{t-1}
=
\frac{1}{\sqrt{\alpha_t}}
\left(
x_t
-
\frac{\beta_t}{\sqrt{1-\bar{\alpha}_t}}
\epsilon_\theta(x_t,t)
\right)
+
\sqrt{\Sigma_\theta(x_t,t)}\,z,
\qquad
z\sim\mathcal{N}(0,I),
\label{eq:reverse_process}
\end{equation}
where \(z\) is Gaussian noise sampled during reverse diffusion. The model is trained using the standard noise prediction objective
\[
\mathcal{L}
=
\mathbb{E}_{t,x_0,\epsilon}
\left[
\|
\epsilon-\epsilon_\theta(x_t,t)
\|_2^2
\right].
\]

\subsection{ControlNet Conditioning}

ControlNet~\cite{zhang-2023} extends pretrained diffusion models by introducing a trainable conditioning branch while keeping the original backbone frozen. The conditioning branch is connected through zero-initialized convolutions, enabling task-specific information to be progressively injected without disrupting the pretrained generative prior. In this work, the conditioning signal corresponds to the event-frame representation introduced above. The output of a conditioned block is given by
\begin{equation}
\bm{y}_c
=
\mathcal{F}(\bm{x};\Theta)
+
\mathcal{Z}
(
\mathcal{F}
(
\bm{x}
+
\mathcal{Z}
(
\bm{c};
\Theta_{z1}
);
\Theta_c
);
\Theta_{z2}
),
\end{equation}
where $\bm{x}$ denotes the input feature map, $\bm{c}$ the conditioning signal (event frame), $\mathcal{F}(\cdot)$ the network block, $\Theta$ and $\Theta_c$ the parameters of the frozen and trainable branches, respectively, and $\mathcal{Z}(\cdot)$ a zero-convolution layer parameterized by $\Theta_{z1}$ and $\Theta_{z2}$.


\section{\method Framework for Handwritten Digit and Face Reconstruction}

\begin{figure}[!h]
    \centering
    \includegraphics[trim=3cm 1cm 0cm 1cm, clip, scale=0.75, width=\linewidth]{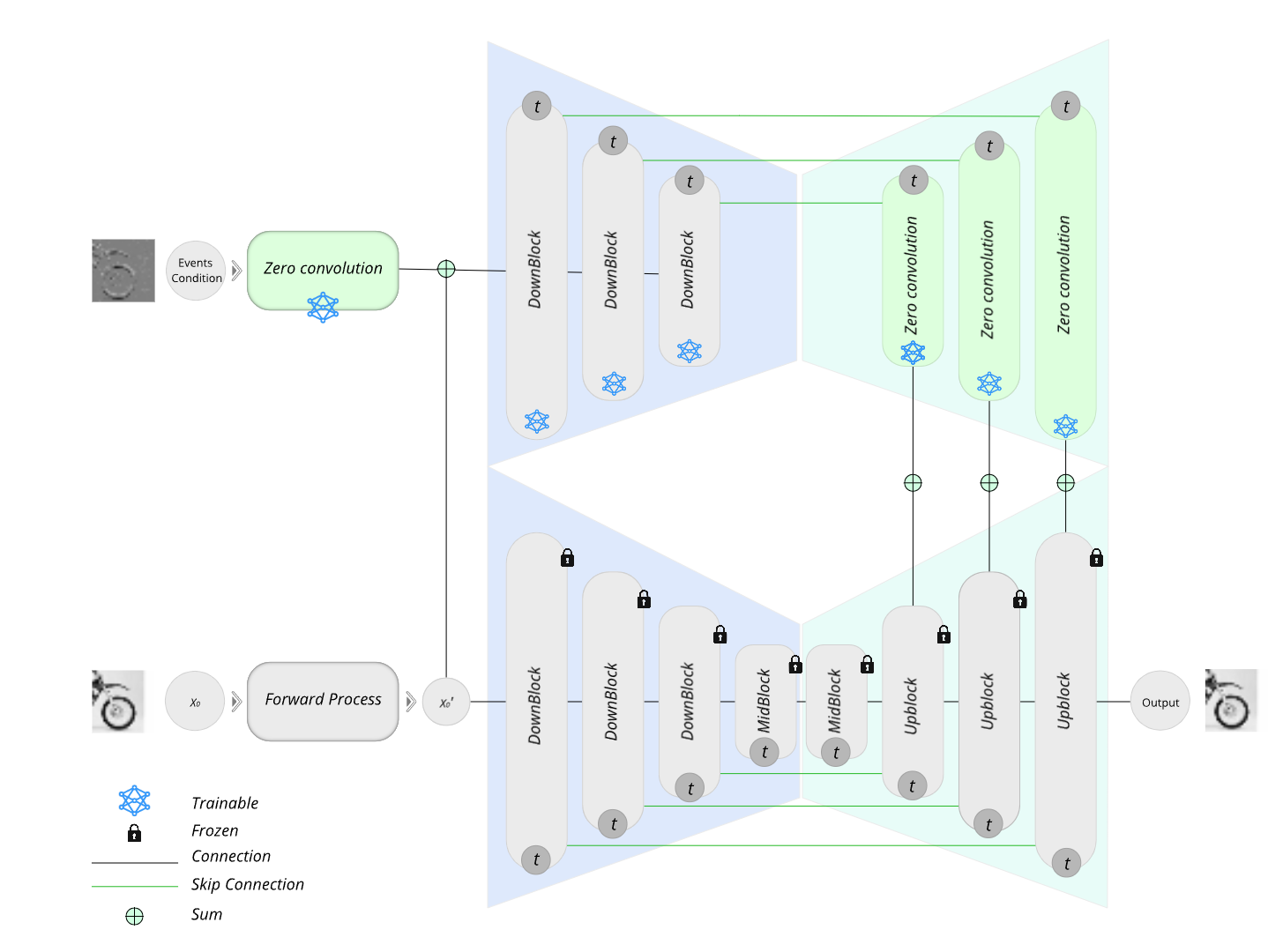}
    \caption{Overview of the proposed \method framework. A pretrained DDPM backbone is conditioned by a trainable ControlNet branch using event frames. Training consists of generic image-prior learning followed by event-guided specialization for the target reconstruction task.}
    \label{fig:ebird-arch}
\end{figure}

The proposed \method framework builds upon DDPMs~\cite{ho2020denoising}, adopting a U-Net backbone~\cite{ronneberger2015} and extending it with ControlNet conditioning~\cite{zhang-2023} for event-guided intensity-image reconstruction. Training follows a two-stage strategy: first, a generic image prior is learned from target images; second, the pretrained backbone is frozen while a trainable ControlNet branch is optimized using event-frame conditioning. Two variants of the proposed framework are considered: \methodg, a generic model trained using all available training samples, and \methods, a specialized model adapted to a particular reconstruction task. The same framework is evaluated on handwritten digit (N-MNIST) and face (RGBE-Gaze) reconstruction. Figure~\ref{fig:ebird-arch} provides an overview of the proposed framework.

\subsection{Framework Overview}

\paragraph{Base Training.}

The first stage learns a generic image prior using only target images, without event information. A noisy sample is generated through the forward diffusion process (Eq.~\ref{eq:forward_diffusion}), and the backbone is optimized to predict the injected Gaussian noise. The pretrained backbone is then frozen for conditional training.

\paragraph{Conditional Training.}

The second stage specializes the pretrained backbone for event-guided reconstruction. The backbone remains frozen, while the ControlNet branch receives the event representation \(c\) and injects conditioning features through zero-convolution layers. The network predicts the injected noise
\(\epsilon_\theta(x_t,t,c)\)
and is optimized using the conditional DDPM objective
\begin{equation}
\mathcal{L}
=
\mathbb{E}_{x_0,t,c,\epsilon}
\left[
\|
\epsilon-\epsilon_\theta(x_t,t,c)
\|_2^2
\right].
\label{eq:ebird_loss}
\end{equation}

\subsection{Network Architecture}

\paragraph{Building Blocks.}

\begin{figure}[!h]
    \centering
    \includegraphics[trim=2cm 8cm 2cm 8cm, clip, scale=0.7, width=\linewidth]{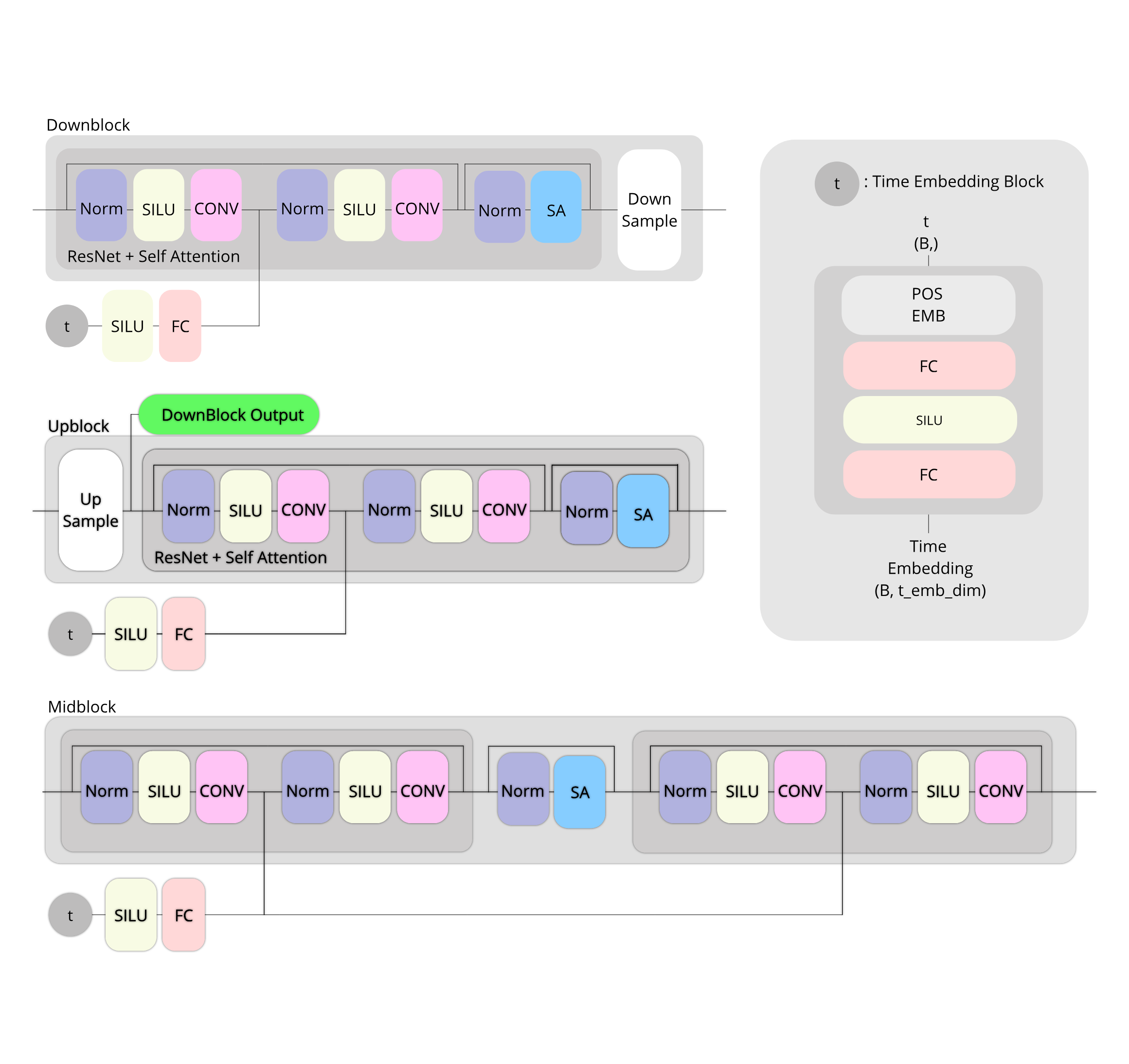}
    \caption{Detailed architecture of the DDPM backbone and ControlNet blocks. Each downsampling, bottleneck, and upsampling module comprises residual convolutional layers with group normalization, SiLU activation, self-attention, and timestep conditioning through sinusoidal positional embeddings.}
    \label{fig:Blocks}
\end{figure}

The proposed framework adopts the standard DDPM U-Net backbone together with a ControlNet branch of identical topology. As illustrated in Fig.~\ref{fig:Blocks}, the network comprises an encoder, a bottleneck, and a decoder connected through skip connections. Each block contains residual convolutions, group normalization, SiLU activations, self-attention, and timestep conditioning. During conditional training, ControlNet injects event-guided features into the frozen backbone through zero-convolution layers at multiple resolutions.

\paragraph{Time Embedding and Activation.}

Each residual block incorporates a sinusoidal time embedding to encode the diffusion timestep. The positional embedding is defined as
\begin{align}
PE_{(t,2i)} = \sin\left(\frac{t}{10000^{2i/d}}\right),
PE_{(t,2i+1)} = \cos\left(\frac{t}{10000^{2i/d}}\right),
\end{align}
where $d$ denotes the embedding dimension. The resulting embedding is injected into every residual block, allowing the network to adapt its denoising behavior throughout the diffusion process. The SiLU activation function
${SiLU}(x)=x\,\sigma(x)$
is used throughout the architecture.


\subsection{Training}

Training follows the two-stage strategy described above. The DDPM backbone is first pretrained using target images and subsequently frozen while ControlNet is optimized using paired event-image samples. Algorithm~\ref{alg:training} summarizes the complete procedure.

\begin{algorithm}[!h]
\caption{Two-Stage Training of eBIRD}
\label{alg:training}
\begin{algorithmic}[1]

\State \textbf{Stage 1: Base Training}
\For{each training iteration}
    \State Sample $x_0$, $t$, and $\epsilon\sim\mathcal{N}(0,I)$
    \State Generate noisy image
    $x_t \leftarrow \sqrt{\bar{\alpha}_t}x_0+\sqrt{1-\bar{\alpha}_t}\epsilon$
    \State Predict $\hat{\epsilon}\leftarrow\epsilon_\theta(x_t,t)$
    \State Update backbone parameters $\theta$ using
    $\mathcal{L}_{\mathrm{base}}=\|\epsilon-\hat{\epsilon}\|_2^2$
\EndFor

\State Initialize ControlNet from the trained backbone and freeze $\theta$

\State \textbf{Stage 2: Task Specialization}
\For{each training iteration}
    \State Sample paired $(x_0,c)$, $t$, and $\epsilon\sim\mathcal{N}(0,I)$
    \State Generate noisy image
    $x_t \leftarrow \sqrt{\bar{\alpha}_t}x_0+\sqrt{1-\bar{\alpha}_t}\epsilon$
    \State Condition the frozen backbone using ControlNet
    \State Predict $\hat{\epsilon}\leftarrow\epsilon_{\theta,\phi}(x_t,t,c)$
    \State Update ControlNet parameters $\phi$ using
    $\mathcal{L}_{\mathrm{cond}}=\|\epsilon-\hat{\epsilon}\|_2^2$
\EndFor

\end{algorithmic}
\end{algorithm}

Two training strategies are considered. \methodg is trained using all available training samples, allowing a single model to reconstruct multiple classes or subjects. In contrast, \methods specializes the pretrained model for a particular reconstruction task. For handwritten digits, specialization is achieved by training an independent model for each digit class. For RGBE-Gaze, \methodg is subsequently fine-tuned using data from the target users.

\subsection{Inference}

During inference, image reconstruction starts from Gaussian noise
$x_T\sim\mathcal{N}(0,I)$ together with an event representation $c$.
At each reverse diffusion step, the frozen DDPM backbone is conditioned by ControlNet to progressively remove noise and reconstruct the target image.
Algorithm~\ref{alg:sampling} summarizes the complete sampling procedure, where the reverse denoising step follows Eq.~(\ref{eq:reverse_process}).
The final denoised sample $x_0$ corresponds to the reconstructed image.

\begin{algorithm}[!h]
\caption{Conditional Sampling Procedure}
\label{alg:sampling}
\begin{algorithmic}[1]
    \State Sample initial noise $x_T \sim \mathcal{N}(0,I)$
    \State Load event representation $c$
    \State Extract conditioning features from $c$ using ControlNet

    \For{$t=T,\dots,1$}
        \State Predict conditional noise
        $\hat{\epsilon}\leftarrow\epsilon_{\theta,\phi}(x_t,t,c)$

        \If{$t>1$}
            \State Sample $z\sim\mathcal{N}(0,I)$
        \Else
            \State Set $z\leftarrow 0$
        \EndIf

        \State Compute the reverse diffusion step
        $
        x_{t-1}
        \leftarrow
        \frac{1}{\sqrt{\alpha_t}}
        \left(
        x_t-
        \frac{\beta_t}{\sqrt{1-\bar{\alpha}_t}}
        \hat{\epsilon}
        \right)
        +
        \sigma_t z
        $
    \EndFor

    \State \Return reconstructed image $x_0$
\end{algorithmic}
\end{algorithm}


\section{Experimental Settings}
\label{sec:experimental-settings}

This section describes the experimental protocol used to evaluate \methodg and \methods on handwritten digit and face reconstruction. We first introduce the datasets and training protocols, followed by the evaluation metrics and hardware configuration adopted throughout the experiments.

\subsection{Datasets}

\paragraph{N-MNIST.}

Originally introduced by Orchard \emph{et al.}~\cite{orchard-2015}, this neuromorphic adaptation of the classic MNIST digit dataset was captured using an ATIS (Asynchronous Time-based Image Sensor), which outputs event sequences encoding spatial location, polarity (brightness increase or decrease), and precise timestamps. This event-based representation enables research in neuromorphic vision and event-driven image reconstruction.
It contains 60,000 training samples ($\sim$6K per digit class) and 10,000 test samples ($\sim$1K per class), preserving the original MNIST spatial resolution of $28\times28$ pixels.
\begin{figure}[!h]
    \centering
    \includegraphics[width=0.09\linewidth]{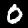}
    \includegraphics[width=0.09\linewidth]{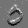}
    \includegraphics[width=0.09\linewidth]{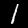}
    \includegraphics[width=0.09\linewidth]{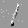}
    \includegraphics[width=0.09\linewidth]{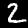}
    \includegraphics[width=0.09\linewidth]{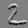}
    \includegraphics[width=0.09\linewidth]{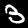}
    \includegraphics[width=0.09\linewidth]{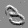}
    \includegraphics[width=0.09\linewidth]{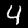}
    \includegraphics[width=0.09\linewidth]{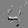}

    \includegraphics[width=0.09\linewidth]{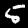}
    \includegraphics[width=0.09\linewidth]{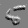}
    \includegraphics[width=0.09\linewidth]{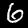}
    \includegraphics[width=0.09\linewidth]{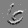}
    \includegraphics[width=0.09\linewidth]{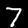}
    \includegraphics[width=0.09\linewidth]{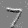}
    \includegraphics[width=0.09\linewidth]{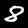}
    \includegraphics[width=0.09\linewidth]{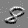}
    \includegraphics[width=0.09\linewidth]{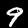}
    \includegraphics[width=0.09\linewidth]{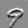}
    
    \caption{Representative samples from the N-MNIST dataset~\cite{orchard-2015}. Each pair consists of the original MNIST grayscale digit (left) and its corresponding event-based representation (right), illustrating the input-target pairs used for handwritten digit reconstruction.}
    \label{fig:nmnist_samples}
\end{figure}

To compare the performance of class-specific and general reconstruction models, we designed 11 experimental scenarios: 10 models trained to reconstruct a single digit class (0--9) and one general model trained to reconstruct all digit classes. All experiments use a temporal window of 33 ms to generate the event representations.

\paragraph{RGBE-Gaze.}

Introduced for high-frequency remote gaze tracking~\cite{zhao2025rgbegaze}, this multimodal dataset provides synchronized full-face RGB images and high-temporal-resolution event streams acquired from 66 participants under diverse head poses, gaze directions, and viewing distances. It comprises approximately 3.6 million RGB images and 26.3 billion event samples captured using a hybrid RGB-event camera setup, providing a large-scale benchmark for event-based vision tasks.

Although originally designed for gaze estimation, this work repurposes RGBE-Gaze as an event-to-image reconstruction benchmark. The synchronized RGB images are used as reconstruction targets, while the corresponding event streams serve as network inputs. Since the objective is to reconstruct facial appearance rather than color information, all RGB target images are converted to grayscale before training and evaluation.
\begin{figure}[!h]
    \centering

    \includegraphics[width=0.2\linewidth]{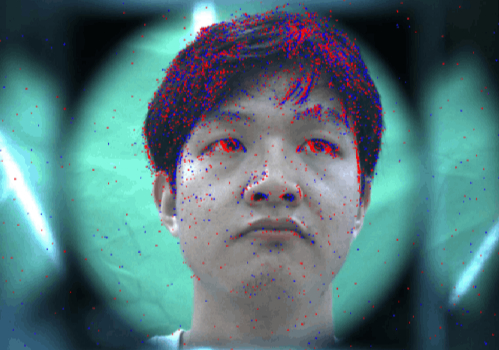}
    \includegraphics[width=0.2\linewidth]{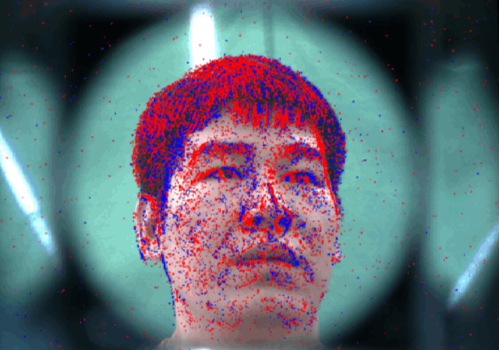}
    \includegraphics[width=0.2\linewidth]{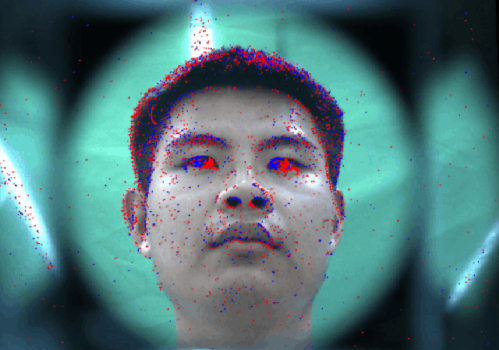}
    \includegraphics[width=0.2\linewidth]{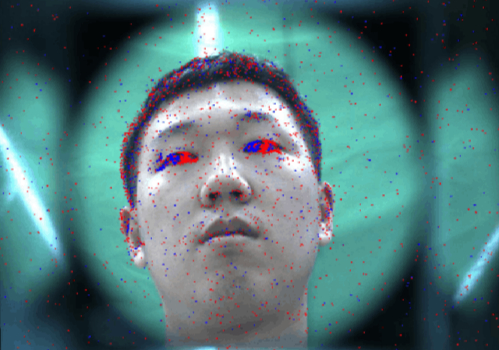}
    
    \includegraphics[width=0.2\linewidth]{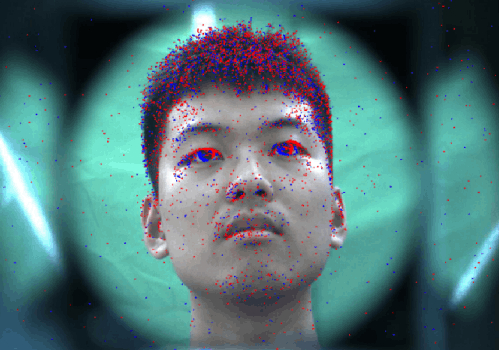}
    \includegraphics[width=0.2\linewidth]{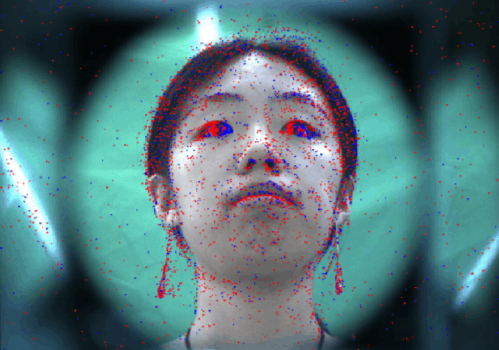}
    \includegraphics[width=0.2\linewidth]{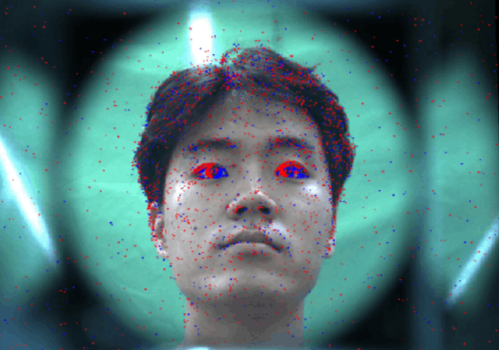}
    \includegraphics[width=0.2\linewidth]{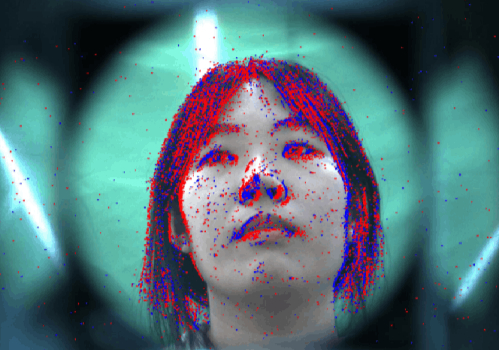}

    \caption{Samples from the RGBE-Gaze dataset~\cite{zhao2025rgbegaze}. Each example shows the RGB target image with the corresponding event stream overlaid. For the experiments presented in this paper, RGB images are converted to grayscale and used as targets.}
    \label{fig:rgbe_samples}
\end{figure}

\subsection{Training and Evaluation Details}

All proposed models were implemented in PyTorch and trained using the Adam optimizer with a learning rate of $1\times10^{-4}$, a batch size of 80, and 40 training epochs. During training, diffusion timesteps were uniformly sampled, and Gaussian noise was added according to the DDPM forward process. The DDPM backbone was first pretrained on target images and subsequently frozen while the ControlNet branch was optimized using paired event-image samples.

Two learning strategies were considered throughout the experiments. \methodg was trained using all available training samples, whereas \methods specialized the pretrained model for the target reconstruction task. For N-MNIST, specialization was achieved by training an independent model for each digit class. For RGBE-Gaze, \methodg was further fine-tuned using data from the target subjects.

For RGBE-Gaze, E2VID~\cite{rebecq2021e2vid} and ET-Net~\cite{weng2021etnet} were evaluated using their publicly available pretrained models without additional fine-tuning. Pix2Pix~\cite{isola2017pix2pix} was trained using the same paired event-image data and optimization settings as the proposed framework. All methods were evaluated using the same train, validation and test split and the MSE, SSIM, and PSNR metrics.

\subsection{Evaluation Metrics}

Reconstruction quality is evaluated using Mean Squared Error (MSE), Structural Similarity Index Measure (SSIM), and Peak Signal-to-Noise Ratio (PSNR).

MSE measures the average pixel-wise reconstruction error between the reconstructed image $\hat{x}$ and the ground-truth image $x$:
$
\mathrm{MSE}=\frac{1}{N}\sum_{i=1}^{N}(x_i-\hat{x}_i)^2,
$
where $N$ is the number of pixels. Lower values indicate higher reconstruction fidelity.

PSNR quantifies reconstruction quality relative to the image dynamic range and is computed from the MSE as
$
\mathrm{PSNR}=10\log_{10}\left(\frac{\mathrm{MAX}^2}{\mathrm{MSE}}\right),
$
where $\mathrm{MAX}$ denotes the maximum possible pixel value. Higher values indicate better reconstruction quality.

SSIM evaluates perceptual similarity by comparing local luminance, contrast, and structural information:
\begin{equation}
\mathrm{SSIM}(x,\hat{x})=
\frac{(2\mu_x\mu_{\hat{x}}+C_1)(2\sigma_{x\hat{x}}+C_2)}
{(\mu_x^2+\mu_{\hat{x}}^2+C_1)(\sigma_x^2+\sigma_{\hat{x}}^2+C_2)},
\end{equation}
where $\mu$, $\sigma^2$, and $\sigma_{x\hat{x}}$ denote the local means, variances, and covariance, respectively. SSIM ranges from $-1$ to $1$, with values closer to 1 indicating greater structural similarity.

\subsection{Hardware}

All experiments were conducted on an NVIDIA DGX-1 system equipped with eight NVIDIA Tesla V100 GPUs, each providing 32~GB of GPU memory. The models were implemented in PyTorch and trained using distributed GPU execution for \methodg and \methods training, validation and evaluation.


\section{Results and Discussion}

This section presents and discusses the experimental results obtained on two complementary event-guided reconstruction tasks. N-MNIST is used to evaluate the impact of generic and class-specific models for handwritten digit reconstruction, whereas RGBE-Gaze extends the analysis to face reconstruction and includes comparisons with sota event-to-image reconstruction methods.

\subsection{Handwritten Digit Reconstruction using N-MNIST}

\begin{figure*}[!h]
    \centering
    \includegraphics[width=\linewidth]{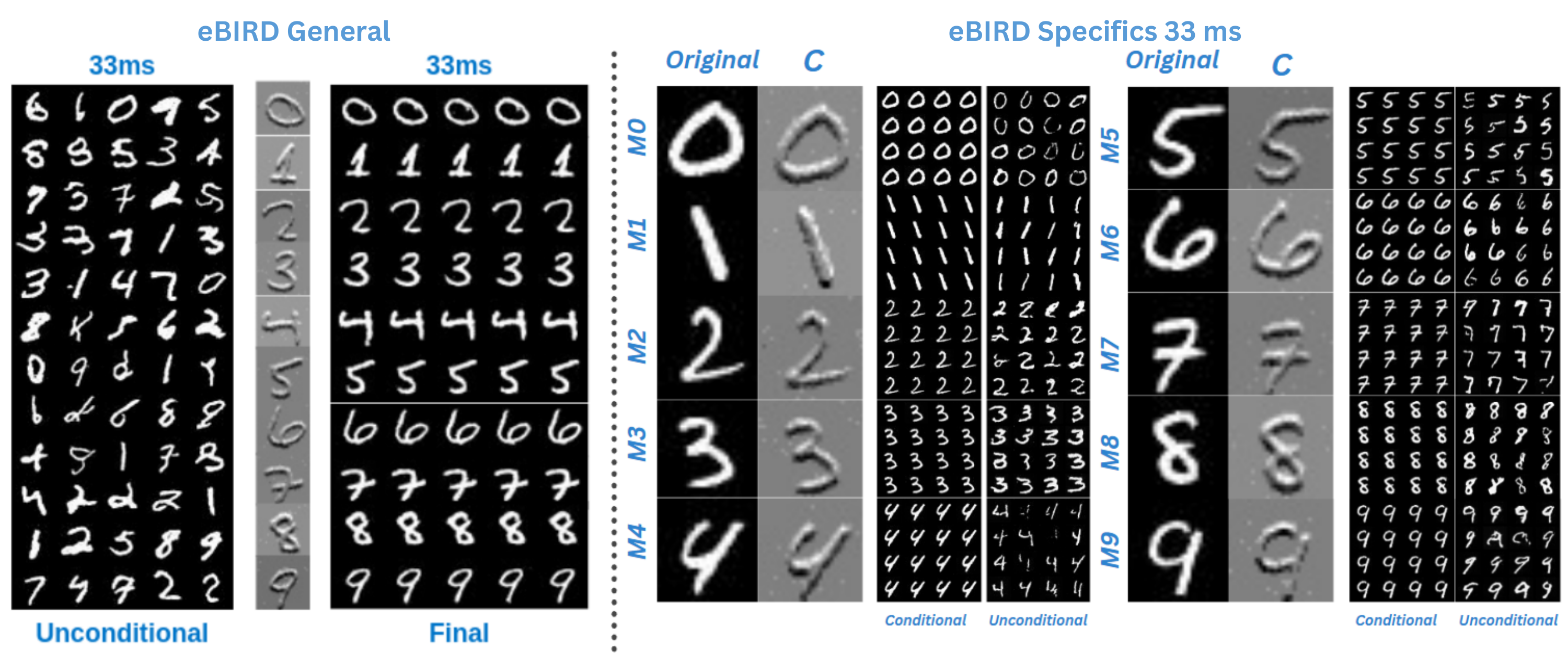}
    \caption{Qualitative reconstruction results on the N-MNIST dataset using 33~ms event windows. The first column shows the event representation, followed by the reconstructions obtained with the class-specific models and the general model, together with the corresponding ground-truth digit. Both approaches successfully reconstruct the handwritten digits, while the general model produces sharper digit boundaries and fewer reconstruction artifacts.}
    \label{fig:nmnist-results}
\end{figure*}

Figure~\ref{fig:nmnist-results} presents representative reconstruction examples obtained with general and class-specific models, while Table~\ref{tab:nmnist-results} reports the corresponding quantitative evaluation using 33~ms event windows.

\begin{table}[!h]
\centering
\caption{Quantitative comparison on the N-MNIST dataset using a temporal window of 33 ms. Lower MSE and higher SSIM and PSNR indicate better reconstruction quality. Class-specific models are trained independently for each digit class, whereas the general model is trained using all digit classes.}
\label{tab:nmnist-results}
\textbf{Class-Specific Models} \\
\vspace{2pt}
    \begin{tabular}{lcccccccccc}
    \toprule
    \textbf{Metric} & \textbf{S1} & \textbf{S2} & \textbf{S3} & \textbf{S4} & \textbf{S5} & \textbf{S6} & \textbf{S7} & \textbf{S8} & \textbf{S9} & \textbf{S0} \\
    \midrule
    MSE   & 0.0044 & 0.0074 & 0.0069 & 0.0075 & 0.0094 & 0.0076 & 0.0074 & 0.0077 & 0.0071 & 0.0082 \\
    SSIM  & 0.8667 & 0.8782 & 0.8788 & 0.8461 & 0.8532 & 0.8764 & 0.8514 & 0.8858 & 0.8594 & 0.8953 \\
    PSNR  & 24.07  & 21.67  & 21.99  & 21.57  & 21.50  & 21.59  & 21.76  & 21.47  & 21.91  & 21.19 \\
    \bottomrule
    \end{tabular}
    
    \vspace{1em}
    
    \textbf{General Model} \\
    \vspace{2pt}
    \begin{tabular}{lcccccccccc}
    \toprule
    \textbf{Metric} & \textbf{G1} & \textbf{G2} & \textbf{G3} & \textbf{G4} & \textbf{G5} & \textbf{G6} & \textbf{G7} & \textbf{G8} & \textbf{G9} & \textbf{G0} \\
    \midrule
    MSE   & 0.0031 & 0.0055 & 0.0056 & 0.0053 & 0.0055 & 0.0054 & 0.0052 & 0.0058 & 0.0053 & 0.0056 \\
    SSIM  & 0.8858 & 0.9059 & 0.8996 & 0.8843 & 0.9014 & 0.9019 & 0.8814 & 0.9067 & 0.8889 & 0.9258 \\
    PSNR  & 25.47  & 23.06  & 23.09  & 23.15  & 23.15  & 23.14  & 23.49  & 22.75  & 23.17  & 22.85 \\
    \bottomrule
    \end{tabular}
\end{table}

Table~\ref{tab:nmnist-results} shows that the general model consistently outperforms class-specific models for each class across all three evaluation metrics. Averaged over the ten digit classes, the general model achieves a lower MSE (0.0052 vs. 0.0074), a higher SSIM (0.8982 vs. 0.8691), and a higher PSNR (23.34 dB vs. 21.88 dB). These results indicate that learning a shared image prior across all handwritten digits is more effective than training independent models for individual classes.

A possible explanation is the relatively limited appearance variability of handwritten digits. Training a single general model allows the diffusion prior to exploit a larger and more diverse set of handwriting patterns while preserving the structural characteristics shared among all classes. Consequently, the general model generalizes more effectively than independently trained specialized models.
Figure~\ref{fig:nmnist-results} qualitatively supports these observations. Both variants successfully reconstruct recognizable handwritten digits from event representations, while the general model generally produces sharper digit boundaries and fewer reconstruction artifacts.

These results motivated evaluating whether the same behavior holds for a more realistic reconstruction task involving natural facial appearance.

\subsection{Face Reconstruction using RGB-E Gaze}

\begin{figure}[!h]
\centering
\subcaptionbox{Input}[0.19\linewidth]{%
    \includegraphics[width=\linewidth]{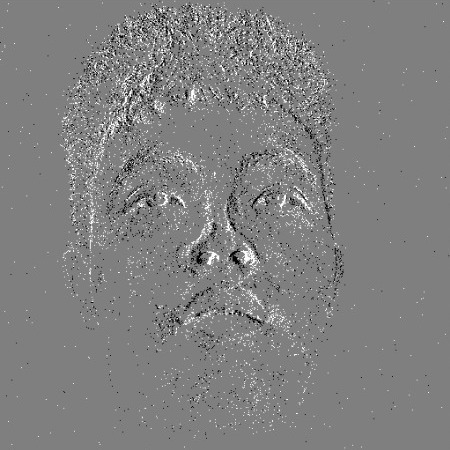}}
\subcaptionbox{Pix2Pix}[0.19\linewidth]{%
    \includegraphics[width=\linewidth]{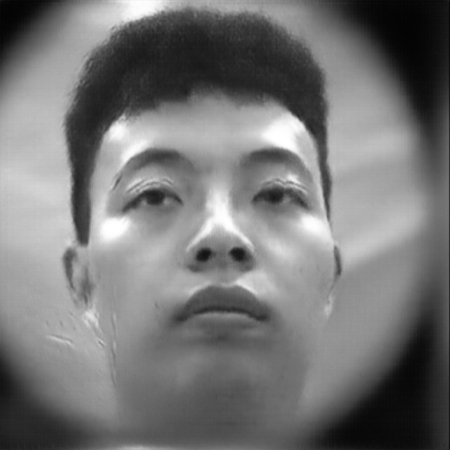}}
\subcaptionbox{\methodg}[0.19\linewidth]{%
    \includegraphics[width=\linewidth]{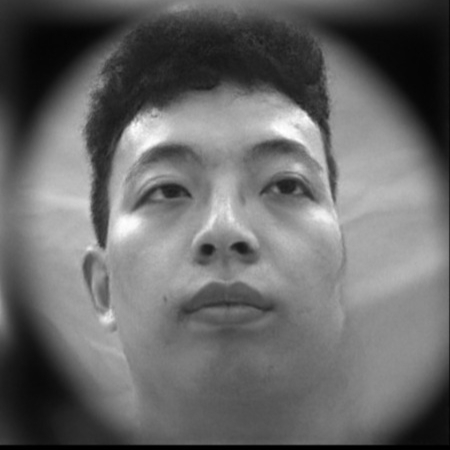}}
\subcaptionbox{\methods}[0.19\linewidth]{%
    \includegraphics[width=\linewidth]{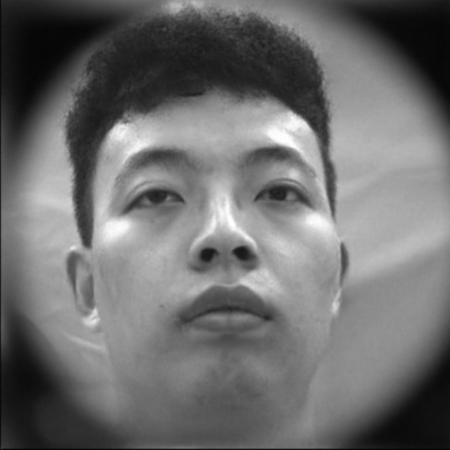}}
\subcaptionbox{Target}[0.19\linewidth]{%
    \includegraphics[width=\linewidth]{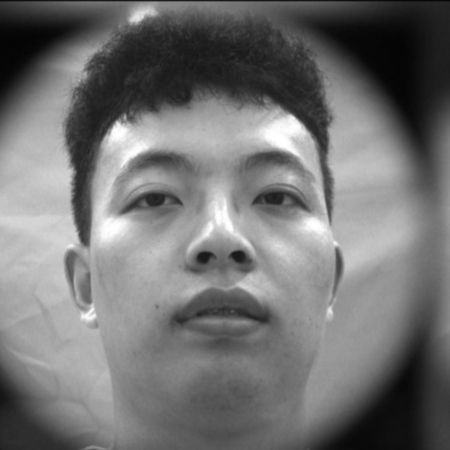}}

\caption{Qualitative comparison of reconstruction methods on the RGBE-Gaze dataset. From left to right: input event representation, Pix2Pix reconstruction, \method generic reconstruction (\methodg), \method specialized reconstruction (\methods), and the corresponding grayscale ground-truth image.}
\label{fig:rgbe-comparison}
\end{figure}

\begin{figure}[!h]
    \centering
    \includegraphics[width=0.47\linewidth]{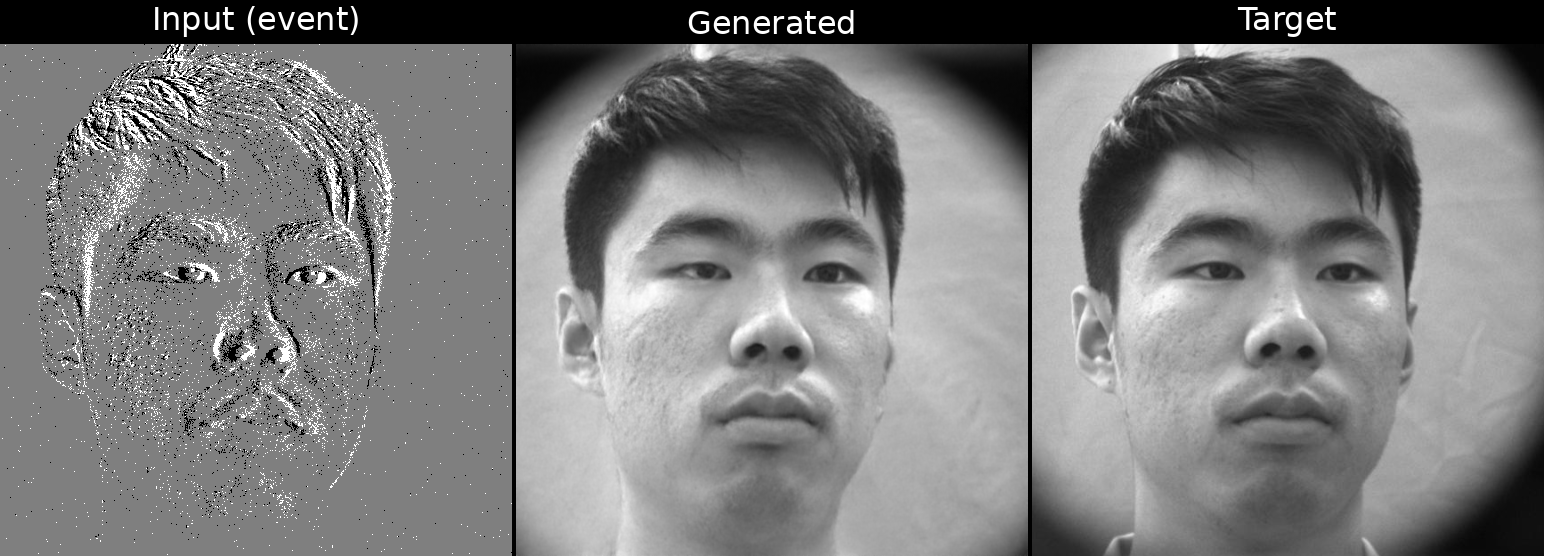}
    \quad
    \includegraphics[width=0.47\linewidth]{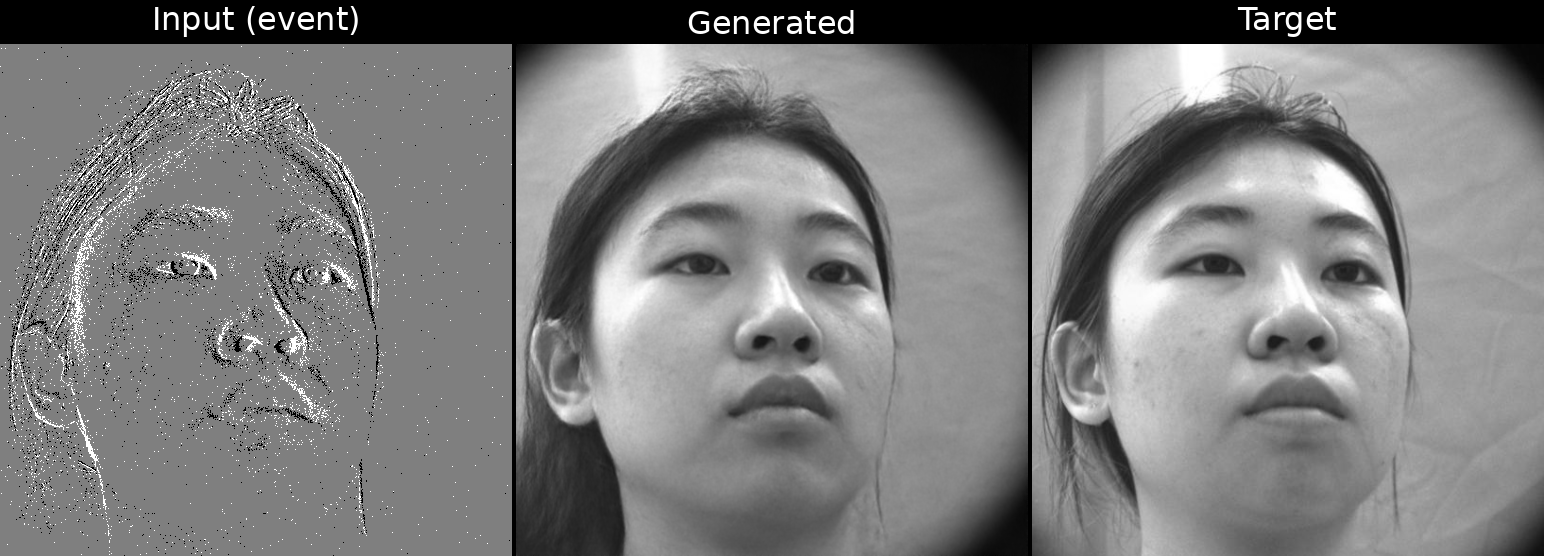}

    \includegraphics[width=0.47\linewidth]{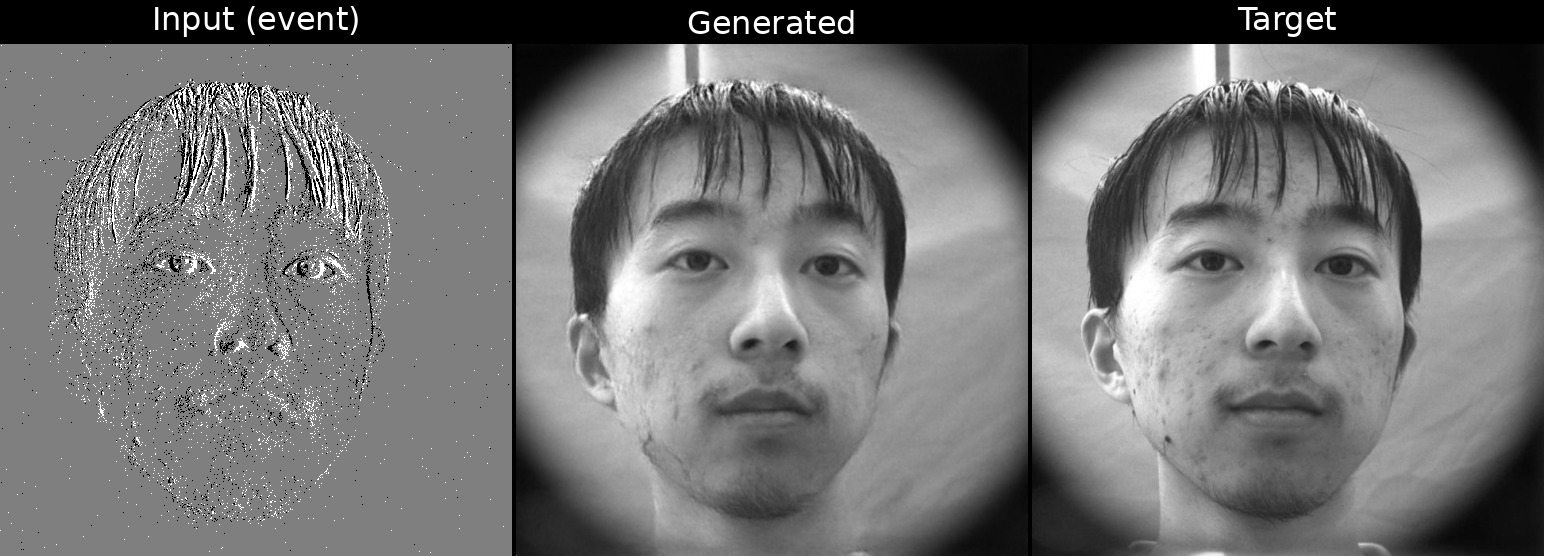}
    \quad
    \includegraphics[width=0.47\linewidth]{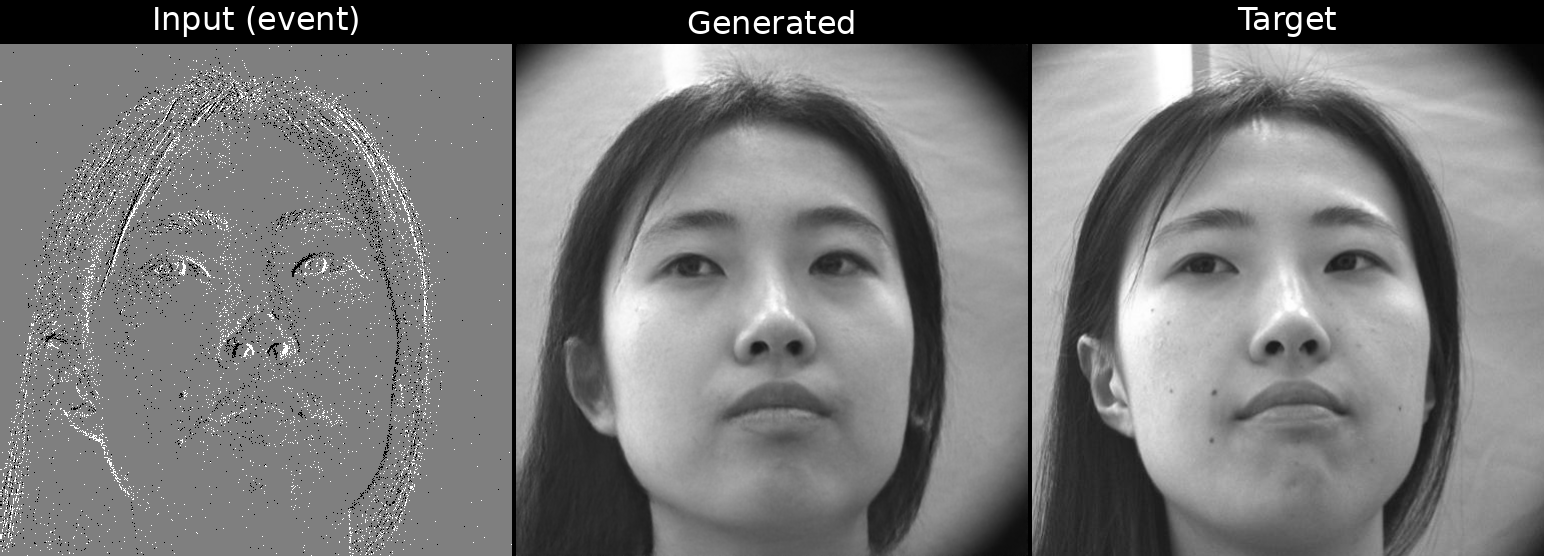}

    \includegraphics[width=0.47\linewidth]{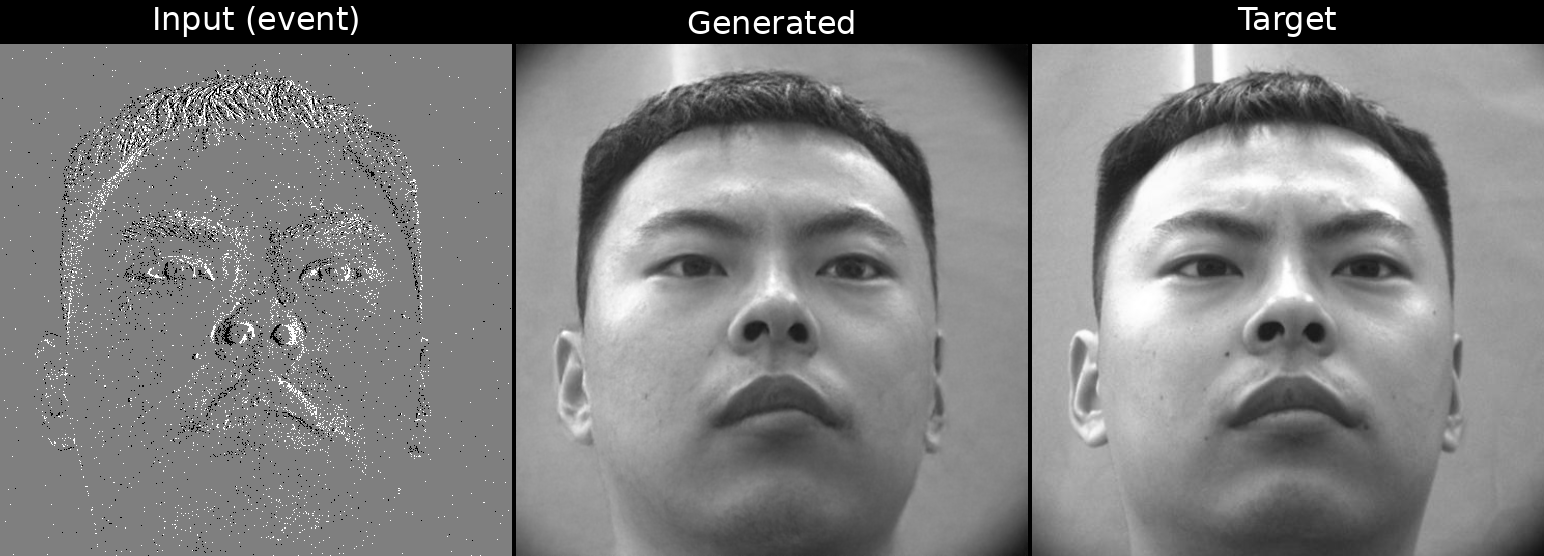}
    \quad
    \includegraphics[width=0.47\linewidth]{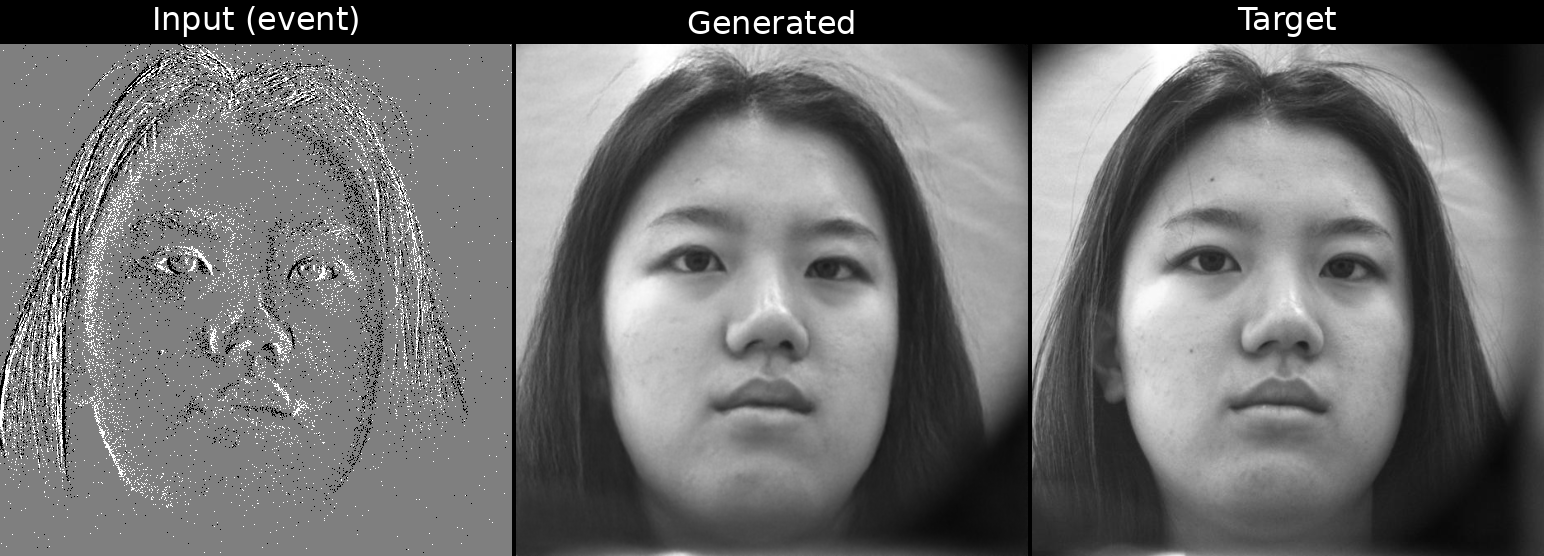}

    \includegraphics[width=0.47\linewidth]{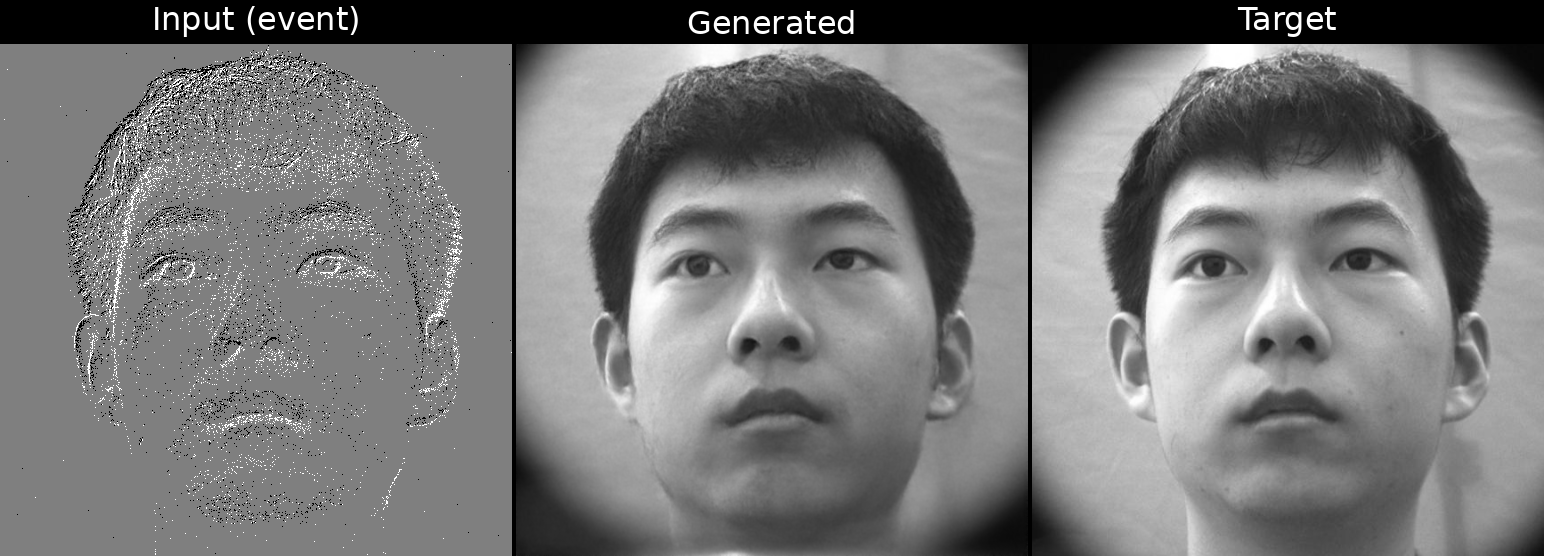}
    \quad
    \includegraphics[width=0.47\linewidth]{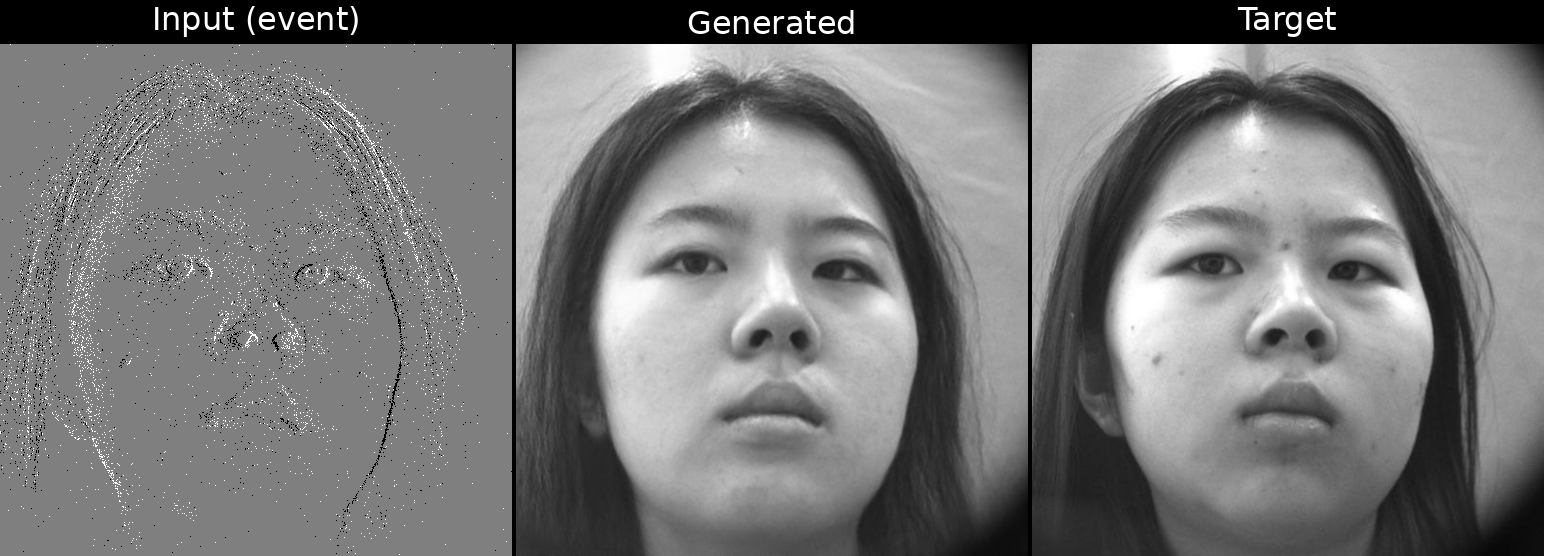}

    \caption{Qualitative reconstruction results on the RGBE-Gaze dataset. Each triplet shows the event representation (left), the image reconstructed by the proposed \method framework (center), and the corresponding grayscale ground-truth image (right). Representative examples from different subjects are shown.}
    \label{fig:rgbe_reconstruction}
\end{figure}

Table~\ref{tab:rgbe_metrics} reports the quantitative evaluation on the RGBE-Gaze dataset, while Figures~\ref{fig:rgbe-comparison} and~\ref{fig:rgbe_reconstruction} present representative qualitative reconstruction results.
\begin{table}[!h]
\centering
\setlength{\tabcolsep}{6pt}
\caption{Quantitative comparison on the RGBE-Gaze dataset. Lower MSE and higher SSIM and PSNR indicate better reconstruction quality. Results are reported as mean $\pm$ standard deviation. The protocol column indicates whether a method is evaluated directly using pretrained weights (E) or trained on RGBE-Gaze (T + E).}
\label{tab:rgbe_metrics}

\begin{tabular}{lcccc}
\toprule
\textbf{Method} &
\textbf{Protocol} &
\textbf{MSE} $\downarrow$ &
\textbf{SSIM} $\uparrow$ &
\textbf{PSNR (dB)} $\uparrow$ \\
\midrule

E2VID~\cite{rebecq2021e2vid}
& E
& $0.1032 \pm 0.02$
& $0.4542 \pm 0.04$
& $9.96 \pm 0.92$ \\

ET-Net~\cite{weng2021etnet}
& E
& $0.0958 \pm 0.02$
& $0.3934 \pm 0.04$
& $10.26 \pm 0.81$ \\

\midrule

\textbf{\methodg (Ours)}
& T + E
& $0.0276 \pm 0.01$
& $0.6381 \pm 0.06$
& $16.15 \pm 2.21$ \\

Pix2Pix~\cite{isola2017pix2pix}
& T + E
& $0.0214 \pm 0.00$
& $0.6714 \pm 0.01$
& $17.49 \pm 0.81$ \\

\textbf{\methods (Ours)}
& T + E
& $\mathbf{0.0161 \pm 0.01}$
& $\mathbf{0.7615 \pm 0.04}$
& $\mathbf{19.08 \pm 3.10}$ \\
\bottomrule
\end{tabular}
\end{table}

Compared with the E2VID and ET-Net recurrent event-to-image reconstruction methods, \method substantially improves reconstruction quality. While E2VID achieves an MSE of 0.1032, an SSIM of 0.4542, and a PSNR of 9.96~dB, and ET-Net obtains an MSE of 0.0958, an SSIM of 0.3934, and a PSNR of 10.26~dB, \methodg reduces the reconstruction error to 0.0276 and improves the SSIM and PSNR to 0.6381 and 16.15~dB, respectively. The specialized variant, \methods, further improves these results, reaching an MSE of 0.0161, an SSIM of 0.7615, and a PSNR of 19.08~dB. These results suggest that combining a pretrained diffusion prior with event-guided conditioning provides substantially higher reconstruction fidelity than recurrent event-to-image reconstruction methods.

Among the trainable approaches, \methods also outperforms the Pix2Pix baseline. Compared with Pix2Pix (MSE 0.0214, SSIM 0.6714, PSNR 17.49~dB), \methods reduces the reconstruction error by approximately 25\%, increases the structural similarity from 0.6714 to 0.7615, and improves the PSNR by approximately 1.6~dB. These improvements indicate that the proposed diffusion framework more accurately preserves facial geometry and appearance while producing fewer reconstruction artifacts.

In contrast to the N-MNIST experiments, \methodg performs worse than \methods. Although \methodg is trained using all available subjects, face reconstruction presents substantially greater variability than handwritten digits due to differences in facial geometry, hairstyle, illumination, expression, and subject appearance. Consequently, fine-tuning the pretrained model for the target subjects enables \methods to better capture identity-specific characteristics and achieve higher reconstruction quality.

Figure~\ref{fig:rgbe-comparison} illustrates these differences for a representative example. Compared with Pix2Pix, \methodg produces smoother intensity transitions and fewer artifacts, while \methods further improves facial contours and local appearance details, yielding reconstructions that are visually closer to the ground-truth image.
Figure~\ref{fig:rgbe_reconstruction} further shows the robustness of \methods across multiple subjects and recording conditions. Despite relying exclusively on event representations, the proposed framework consistently reconstructs the global facial geometry and preserves the principal identity-related characteristics under different poses and facial appearances.

Taken together, the N-MNIST and RGBE-Gaze experiments suggest that the preferred learning strategy depends on the variability of the reconstruction task. Generic learning is advantageous when the target domain exhibits limited intra-class variability, whereas specialized adaptation provides superior reconstruction quality for more diverse visual domains such as human faces. Moreover, the results indicate that the proposed framework is applicable beyond simple handwritten digit reconstruction, extending to more realistic event-guided reconstruction tasks involving natural facial appearance, larger intra-class variability, and challenging visual conditions.

\section{Conclusions and Future Work}

This work presented \method, an event-based intensity-image reconstruction framework that combines a DDPM with ControlNet conditioning to reconstruct intensity images from event representations. The proposed two-stage training framework separates the learning of a generic image prior from event-guided conditioning, enabling effective adaptation to different event-based reconstruction problems.

Experimental results on the N-MNIST and RGBE-Gaze datasets suggest that the proposed framework achieves consistent reconstruction performance across two complementary domains. For handwritten digit reconstruction, a general model trained across all digit classes consistently outperformed independently trained class-specific models. In contrast, for face reconstruction, the specialized adaptation implemented in \methods achieved the best performance, outperforming both the generic variant (\methodg) and representative event-to-image reconstruction methods, including E2VID and ET-Net. These findings suggest that model specialization should be viewed as a property of the reconstruction domain rather than of the reconstruction architecture itself.

Future work will include a comprehensive ablation study of the proposed framework and learning strategies, exploration of alternative event representations to further improve reconstruction quality and generalization, and evaluations on more challenging event-based datasets involving deformable objects and dynamic scenes.

\section*{Acknowledgment}
This work was partially supported by the FONDEQUIP Project EQM170041.

%
%
\bibliographystyle{splncs04}
\bibliography{main}

\end{document}